\documentclass{ar-1col-S2O}
\usepackage[numbers]{natbib}
\usepackage{url}
\usepackage[T1]{fontenc}

\newcommand{\revision}[1]{\textcolor{black}{{{{#1}}}}}

\jname{Xxxx. Xxx. Xxx. Xxx.}
\jvol{AA}
\jyear{YYYY}
\doi{10.1146/((please add article doi))}

\begin{document}

\markboth{Bi et al.}{Machine Learning in Robotic Ultrasound Imaging}

\title{Machine Learning in Robotic Ultrasound Imaging: Challenges and Perspectives}






\author{Yuan Bi*, Zhongliang Jiang*, Felix Duelmer, Dianye Huang, and Nassir Navab
\affil{* The first two authors contributed equally. \\
Computer Aided Medical Procedures, Technical University of Munich, Munich, Germany; \\
email: yuan.bi@tum.de, zl.jiang@tum.de, felix.duelmer@tum.de, dianye.huang@tum.de, nassir.navab@tum.de}
}

\begin{abstract}
This article reviews the recent advances in intelligent robotic ultrasound (US) imaging systems. We commence by presenting the commonly employed robotic mechanisms and control techniques in robotic US imaging, along with their clinical applications. Subsequently, we focus on the deployment of machine learning techniques in the development of robotic sonographers, emphasizing crucial developments aimed at enhancing the intelligence of these systems. The methods for achieving autonomous action reasoning are categorized into two sets of approaches: those relying on implicit environmental data interpretation and those using explicit interpretation. Throughout this exploration, we also discuss practical challenges, including those related to the scarcity of medical data, the need for a deeper understanding of the physical aspects involved, and effective data representation approaches. Moreover, we conclude by highlighting the open problems in the field and analyzing different possible perspectives on how the community could move forward in this research area.

\end{abstract}

\maketitle


\section{INTRODUCTION}


\par

Over the past few decades, particularly in the last ten years, the advancement of autonomous medical robots has attracted increasing attention~\cite{dupont2021decade, yang2017medical}. An intelligent robotic colleague is envisioned to work with medical staffs in hospitals owing to the recent advances in fundamental sensing systems and artificial intelligence~\cite{yang2020combating, yip2023artificial, zemmar2020rise}. 
Among the various subfields of medical robotics, robotic sonography has particularly garnered attention from both the scientific and industrial sectors~\cite{jiang2023robotic,von2021medical,li2021overview}. 
This surge in interest can be attributed to the fact that robotic ultrasound (US) examinations are generally non-invasive compared to other surgical robots, resulting in fewer ethical, legal, and regulatory concerns.

\par
\begin{figure}[h]
\includegraphics[width=0.98\textwidth]{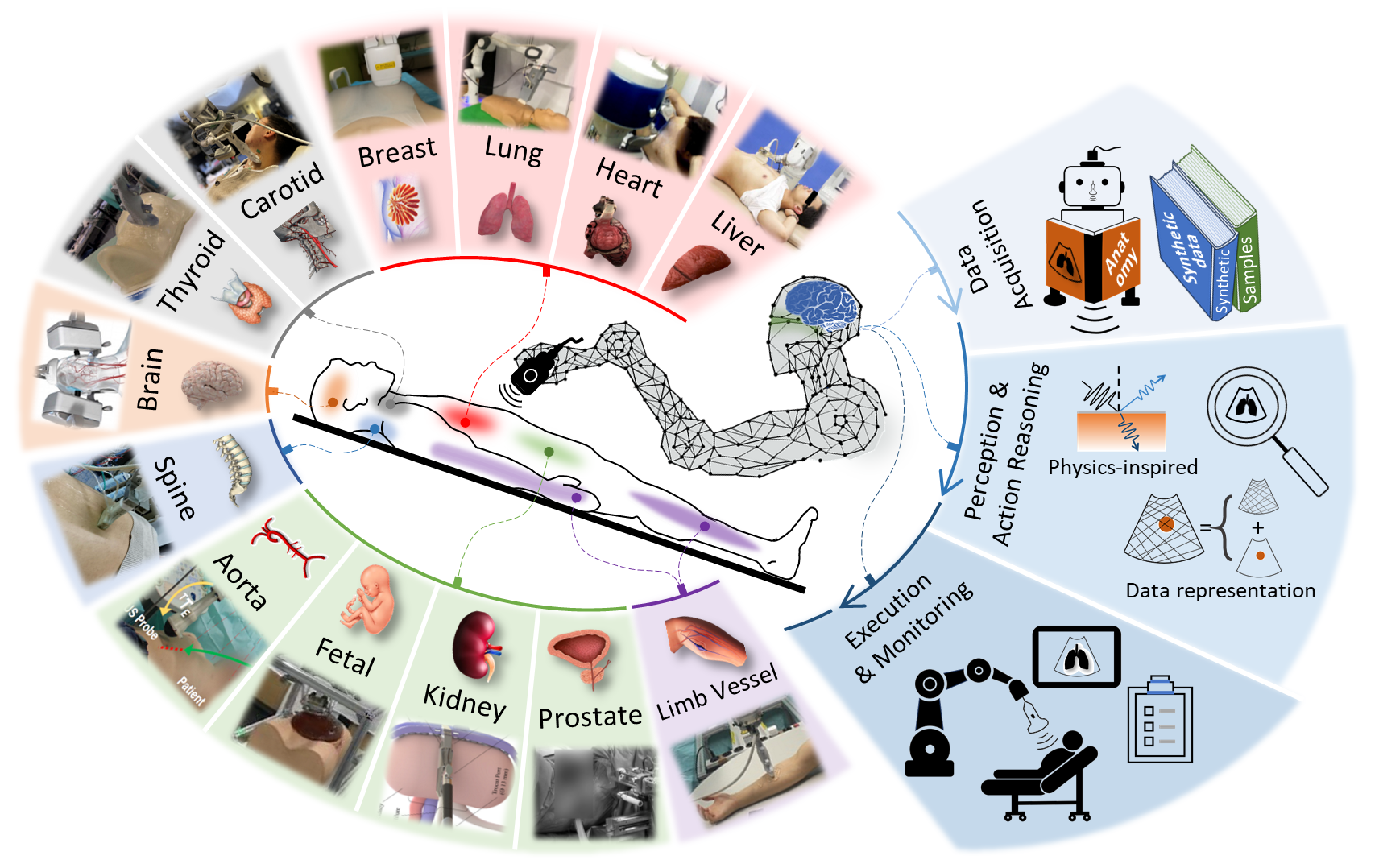}
\caption{Robotic ultrasound imaging systems. 
Left: Typical clinical applications in robotic US imaging. Starting at 2 o'clock and proceeding counter-clockwise: liver~\cite{mustafa2013development}, heart~\cite{giuliani2020user}, lung~\cite{ma2021autonomous}, breast~\cite{tan2022flexible}, carotid~\cite{huang2023motion}, thyroid~\cite{zielke2022rsv}, brain~\cite{esmaeeli2020robotically}, spine~\cite{tirindelli2020force}, aorta~\cite{virga2016automatic}, fetal~\cite{shida2021heart}, kidney~\cite{stilli2019pneumatically}, prostate~\cite{hungr20123}, limb vessels~\cite{jiang2022towards}.
Right: To autonomously maneuver a US probe toward diagnostic views, an intelligent robotic sonographer is envisioned to be capable of leveraging prior anatomical knowledge and analyzing real-time observations for action reasoning.
}
\label{fig:overviewDiagram}
\end{figure}

Due to the advantages of being portable, real-time, and ionizing radiation-free, medical US has gained widespread use in primary healthcare. The rapid evolution of medical imaging modalities throughout the last century has transformed imaging into an indispensable element of standard screening and diagnostic protocols. 
However, the development of interactive and dynamic imaging for treatment guidance and intervention has lagged behind. 
This can be attributed, in part, to the intricate patient and procedure-specific demands they entail. Moreover, their considerable reliance on user proficiency and usability considerations has also contributed to this gradual advancement~\cite{navab2016personalized}.
To improve reproducibility and ensure consistent diagnosis, Salcudean~\emph{et al.} introduced the use of robotic systems to assist in US acquisition~\cite{salcudean1999robot}. By accurately maneuvering the US probe, robotic US systems (RUSS) are expected to standardize examination protocols and optimize the imaging quality~\cite{jiang2020automatic, jiang2020automatic_TIE}. In light of the increasing demand for healthcare interventions and the uneven distribution of experienced sonographers, there is a pressing need for the development of RUSS systems with a high degree of autonomy. \revision{The applications for representative clinical scenarios are summarized in Fig.~\ref{fig:overviewDiagram}.}


\par
Robotic systems have demonstrated the potential to surpass human capabilities in certain domains. Advanced mechanical and material designs, coupled with innovative control methods, enable robots to operate remotely in challenging environments, such as the lunar surface~\cite{zhang2023robotic} and the deep sea~\cite{li2021self}. Nevertheless, there is still a big gap between robotic systems and humans regarding their capabilities for deep understanding and modeling of the dynamic world. Similar to their human counterparts, intelligent robotic sonographers should possess the ability to comprehend human anatomy, diseases, physiology, as well as the physics of US imaging. 
Only by possessing these capabilities, a robotic sonographer can reason for proper actions while performing US examinations and, at the same time, taking the utmost consideration for patient safety.
The two key abilities in this context are perception and execution, which are tightly entangled and can mutually boost each other. The precise scene understanding will benefit the action reasoning process, while the accurate maneuvering of the US probe will further facilitate the understanding. Unfortunately, such intelligent systems still have yet to emerge within the community. But thanks to the boom of machine learning and deep learning, researchers are approaching this ultimate goal from various perspectives. Jiang~\emph{et al.} introduced the concept of ``the Language of Sonography" by inferring the underlying reward function from a few US examination demonstrations from experts to understand the intention of the sonographer~\cite{jiang2023intelligent}. Baumgartner~\emph{et al.} proposed SonoNet, a deep Convolutional Neural Network (CNN), to automatically detect 13 fetal standard planes in real-time~\cite{baumgartner2017sononet}. In addition, Droste~\emph{et al.} presented a
behavior cloning framework to mimic sonographers to search for standard planes~\cite{droste2020automatic}.

\par
Several survey articles have investigated RUSS from different perspectives. For instance, Jiang~\emph{et al} discussed the advancements in terms of complexity of technologies~\cite{jiang2023robotic}, while others categorized papers based on autonomy levels~\cite{von2021medical, li2021overview}. In contrast, this survey article uniquely focuses on the technologies aimed at enhancing the intelligence of RUSS. \revision{Section~\ref{Sec:II_harware} presents an overview of the hardware and control algorithms in RUSS.}
Then, we highlight studies using machine learning technologies to improve the understanding of complex natural and physiological scenes or to directly reason proper actions based on implicit scene comprehension in Section~\ref{sec:III_machine_learning}. Additionally, we have examined potential solutions to tackle practical challenges, including the shortage of medical data, the necessity for considering physical aspects, and the implementation of effective data representation methods in Section~\ref{sec:IV_advanced_intelligence}. Finally, we conclude by presenting several perspectives for the future of intelligent RUSS.

\section{Hardware and Control in Robotic Ultrasound}~\label{Sec:II_harware}
To acquire high-quality US images for diagnostic or analytical purposes, the RUSS should be able to maneuver the US probe towards the target region while precisely regulating several acquisition parameters~\cite{jiang2020automatic}. These parameters primarily involve contact forces, positions, and orientations. The desired values are either determined from the clinician's side within the teleoperated/semi-autonomous paradigms~\cite{salcudean1999robot, li2022dual}, or derived from intuitive/intelligent algorithms~\cite{jiang2020automatic_TIE, chatelain2017confidence} running in the background. Consequently, the evolution of RUSS's mechanism design and control strategy is prominently centered around the enhancement in imaging quality, control accuracy, and execution safety. To highlight the main focus of applying machine learning to RUSS, this section will only provide a brief summary of relevant hardware and control methods used for developing RUSS. A more systematic summary can be referred to a previous survey paper~\cite{jiang2023robotic}. 

\subsection{Mechanism Design}
Safety stands as the highest priority in the development of RUSS. The implementation of specially designed mechanisms can passively limit the applied forces on the patient and avoid potential failure of the sensor or the electrical system, thereby bolstering safety throughout US scanning.
To this end, Tsumura~\emph{et al.} developed a hardware platform to position a US probe for obstetric examination~\cite{tsumura2020robotic}. The platform includes a linear actuator to limit the contact force and a ring-guide mechanism to passively adjust the US probe's posture relative to the body surface. To ensure safety at the actuator level, Wang~\emph{et al.} investigated a customized spring-loaded ball clutch joint to passively restrict the exerted force~\cite{wang2019analysis}. 

\par
Considering quick deployment and fast adaptation to different scanning tasks, it is also common to use general robotic manipulators to develop RUSS~\cite{mustafa2013development, welleweerd2020automated}.
In this case, the mechanism design focuses more on refining the US probe holder component. Tan~\emph{et al.} proposed a dual arm system with a spring-based flexible US probe clamping device~\cite{tan2022flexible} for breast US imaging. In order to passively adapt to the irregular contact contours, Flores~\emph{et al.} adopted a soft probe, which provides good surface adaptation and force transfer, as part of the RUSS system~\cite{facundo2020design}. Wang~\emph{et al.} developed a compliant joint-based adjusting mechanism attached to the robot manipulator to facilitate spine US examinations~\cite{wang2023compliant}. Recently, a motorized probe holder for a robot manipulator has been reported. Bao~\emph{et al.} designed a motor-spring-based mechanical US probe holder to realize the online force adjustment and the operation continuity~\cite{bao2023novel}.
This design pattern can be a promising way to achieve compliant contact for safety and active posture alignment between the probe and contact surface for better image quality.

\subsection{Control Method}\label{sec:controlRUSS}

\par
In the context of US scanning, the full control of RUSS can be decoupled into two subspaces: one degree-of-freedom (DOF) force control along the imaging direction of the probe, and the other five-DOF motion control for navigation along a scanning path~\cite{jiang2023robotic}. To orient an US probe, velocity-based hybrid force-motion control schemes are commonly used in the community~\cite{goel2022autonomous, napoli2018hybrid}. Considering US examinations requiring extensive contact with patients, impedance/admittance control is frequently employed to dynamically modulate contact force and position, primarily for safety concerns~\cite{li2022dual, dyck2022impedance, fang2017force}. 
In order to suppress the accumulated control error, Wang~\emph{et al.} proposed a practical full-task-space compliant control framework with motion estimation and feedback re-initialization for RUSS~\cite{wang2023task}. Nevertheless, the performance of impedance/admittance control heavily relies on the precise estimation of the environmental parameters, specifically stiffness and damping terms. 

\par
Due to the inter-patient variation of tissue properties, varying levels of force are often required to result in high-quality images~\cite{virga2016automatic, guerrero2003deep}. 
To accommodate for changes in contact conditions, Duan~\emph{et al.} proposed a quadratic-programming-based control framework with variable impedance for US-guided scoliosis assessment~\cite{duan2022ultrasound}. Xiao~\emph{et al.} employed variable impedance control to regulate the contact force for better visualization in a human-in-loop environment~\cite{xiao2021visual}. To effectively govern the contact dynamics, Santos~\emph{et al.} proposed a hierarchical control structure where the Cartesian force control takes the primary priority and orientation is controlled within the null space~\cite{santos2018computed}. Their work incorporated online estimation of contact stiffness, enabling control adaptation before contact. Alternatively, Kim~\emph{et al.} utilized the Hunt-Crossley model to formulate a control law with online parameter estimation~\cite{kim2019gain}. Abbas~\emph{et al.} devised an adaptive event-triggered motion-force control scheme for transversal abdomen scan~\cite{abbas2021event}. The control scheme combines an adaptive backstepping motion controller with a PID force controller, and the update of control input is triggered based on Lyapunove-based analysis.\revision{
In order to} integrate RUSS into real-world scenarios, various efforts have been made to optimize the acquisition process using multi-sensor feedback. These optimizations include adjusting US probe's posture~\cite{jiang2020automatic, jiang2020automatic_TIE, ma2021autonomous, wang2022full, raina2023deep}, adapting contact forces~\cite{xiao2021visual, akbari2021robotic}, refining scanning trajectories~\cite{goel2022autonomous, sutedjo2022acoustic}, etc.

\par
\revision{
The integration of robotic systems has showcased the capacity to elevate the precision and consistency of US image acquisition. We consider that the soft mechanics and control techniques in RUSS are relatively well developed, and the key challenge lies in comprehending dynamic environments and action reasoning – a crucial hurdle on the path to realizing intelligent robotic sonographers. In the following sections, we will provide some insights into the application of machine learning algorithms that try to address this challenge.}

\section{Machine Learning in Robotic Ultrasound}~\label{sec:III_machine_learning}
\begin{figure}[h]
\includegraphics[width=0.95\textwidth]{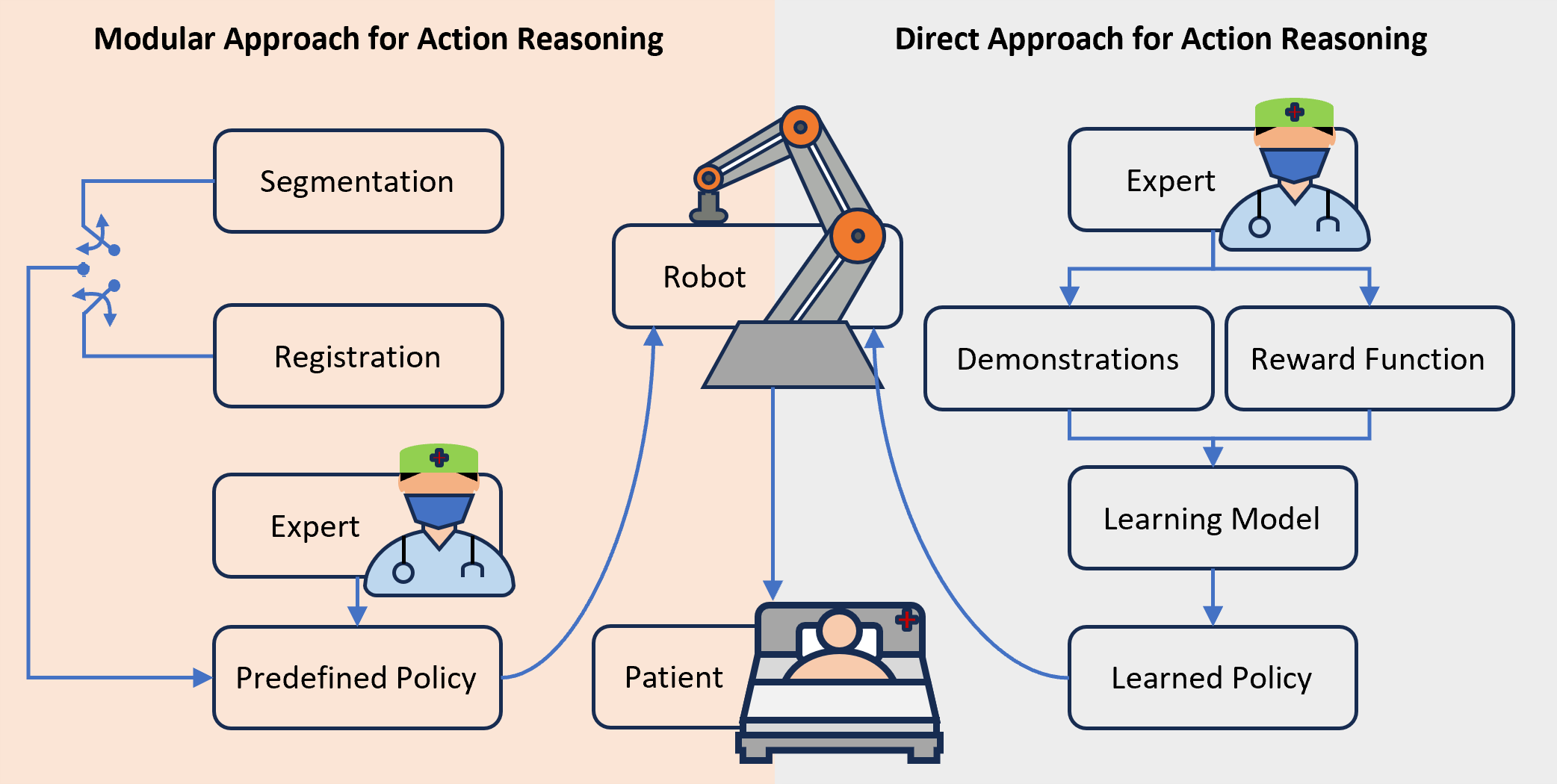}
\caption{Explanation of modular and direct approaches for action reasoning. For modular approaches, the control strategy is defined by a human expert based on different intermediate results provided by machine learning models. Direct approaches refer to machine learning approaches, where the action policy is directly learned from demonstrations or interactions. }
\label{fig:MLbasedRUSS}
\end{figure}

In this section, the machine learning approaches in RUSS are categorized into modular and direct approaches for action reasoning \revision{(Fig.~\ref{fig:MLbasedRUSS})}.
Direct approaches refer to advanced learning frameworks that continuously update and optimize for the best control strategy by directly observing the environment and output corresponding actions.
On the contrary, modular approaches mainly include control algorithms that integrate the outcomes of deep learning models into rule-based control laws. In modular approaches, intermediate perception results are explicitly extracted, while for direct approaches, the intermediate clues are implicitly processed inside a decision-making network. \revision{Both approaches have their own advantages and disadvantages. The direct approach requires less supervision from experts, while the modular approach is more predictable.}

\subsection{Modular Approach for Action Reasoning}
\revision{For human sonographers, the action reasoning is deduced based on their understanding of US image formation in US examinations. In modular approaches, machine learning algorithms explicitly model these implicit interpretation processes of sonographers. Two of the most commonly used analysis techniques are semantic segmentation and registration. The former provides the morphological and positional information of the target anatomy, while the latter offers the feasibility of leveraging pre-operative or generic models. Both techniques will be discussed in detail in this section.}

\subsubsection{Segmentation}\label{sec:Segmentation}
Semantic segmentation of the US image is a commonly used method in RUSS to specify and localize the target anatomy. In practice, the semantic segmentation process is also frequently carried out implicitly in the sonographers' mind during US examinations. To enable autonomous tracking of the target tissue, the outcomes of the segmentation process are incorporated into the robot's control loop. Such frameworks have been widely implemented for different US scanning tasks. Merouche~\emph{et al.} implemented a fast-marching segmentation method to locate the vessel lumen, subsequently directing the robot to maintain the vessel at the center of the US view~\cite{merouche2015robotic}. Similarly, Akbari~\emph{et al.} segmented seroma in breast US using a threshold method to provide feedback to the robot control loop~\cite{akbari2021robot}. 

\par
However, such conventional segmentation methods often lack the intuitive understanding of the anatomic information because they rely on gray level probability density functions. Therefore, the performance of these methods often suffers from noise, speckle, and artifacts.
Recently, deep CNNs have emerged as a powerful alternative to conventional segmentation methods, demonstrating phenomenal success in medical image analysis~\cite{liu2019deep}. In general, there are two characteristics that need to be properly considered: \emph{contour} and \emph{texture}. Contour refers to the statistical shape model of the target anatomy, providing a robust cue for the model during segmentation tasks. However, not all anatomies exhibit stable statistical shapes, especially in cases involving lesions or abnormalities. Hence, the model should also be able to distinguish the segmentation targets based on their unique textures. In the context of CNN, anatomy contour extraction depends largely on the network's capability to address large-scale relationships within the image, whereas the efficiency of texture recognition is closely linked to the network's performance in perceiving local details of images. 

\par
Currently, one of the most popular and effective network architectures for biomedical image segmentation is the U-shaped network~\cite{ronneberger2015u}. Its success is mainly attributed to the skip connections at different scales of the encoder-decoder architectures. Such implementation ensures that the network can take advantage of fine details provided by high-level features and global information extracted from low-level features.
Based on a real-time segmentation of tubular structures using U-Net, Jiang~\emph{et al.} proposed a framework to simultaneously compute the radius of the vessel and optimize the vascular centerline to determine the scanning path~\cite{jiang2021autonomous}. Koskinopoulou~\emph{et al.} designed a dual robot system for venous intervention guided by a segmentation of the veins using Mask R-CNN\revision{, which is a segmentation network designed for simultaneously detection and segmentation of objects}~\cite{koskinopoulou2023dual}. The aforementioned two RUSS have been validated on phantoms. 
To further validate the practical applicability of autonomous RUSS in real-world conditions, Chen~\emph{et al.} integrated an external camera with US vessel detection outcomes from U-Net. This enabled automatic US scanning and subsequent reconstruction of the patient's femoral artery.~\cite{chen2023fully}
An autonomous venous puncture mechanism was designed by Chen~\emph{et al.}, where a semi-supervised segmentation network, semi-ResNeXt-Unet, was presented to address the problem of insufficient labels~\cite{chen2021semi}. The accuracy of the vein segmentation was tested on volunteers and later used to guide the puncture robot. To improve the stability of artery segmentation in US images, Jiang~\emph{et al.} introduced DopUS-Net, which combined Doppler images with B-mode images~\cite{jiang2023dopus}. During the autonomous scanning of radial arteries, the presence of the Doppler effect is constantly monitored, and an out-of-plane fan-motion of the US probe is executed when necessary to ensure good segmentation. Doppler images were also employed by Ning~\emph{et al.} to serve as a weak supervision for the segmentation network~\cite{ning2023autonomous}. 

\par
However, it is worth noting that U-Net and its variants are not specifically designed for US images. The inter-patient or even inter-machine differences of US images are relatively larger than those observed in optical or CT images. As a result, distinct US-oriented learning approaches are necessary to pave the way toward stable and reliable segmentation performance. In this regard, the recent works that explicitly integrated US physics into learning models, e.g., CACTUSS~\cite{velikova2022cactuss}, and LOTUS~\cite{velikova2023lotus}, appear to be a promising solution, which is discussed in detail in the following section, Sec.~\ref{sec:physics-inspired}.

\subsubsection{Registration}
\par
Registration is a commonly employed way for RUSS to map an unseen patient-specific scenario to a generic template, preoperative CT, or MRI. This is, in fact, what human sonographers do regularly. They use their full understanding of human anatomy acquired throughout their education and training to explore the anatomy of interest and navigate to required US views for a given clinical task. For robotic sonographers, MRI or CT atlases could play the role of learned mental mapping. 

\par
It should be noted that this section will only focus on the application of the registration approach on RUSS rather than a comprehensive summary of technical details about registration. For the later purpose, please refer other articles such as~\cite{che2017ultrasound}. Registration techniques have often been used for path planning in developing autonomous RUSS, particularly for the ones from preoperative CT or MRI to the current setup. To this end, Hennersperger~\emph{et al.} employed a classic iterative closest point (ICP) algorithm to align the patient surface extracted from a live RGB-D camera to preoperative MRI~\cite{hennersperger2016towards}. Langsch~\emph{et al.} performed a rigid registration between US volume and MRI/CT volume using $LC^{2}$ similarity metric to determine the scanning path for aorta~\cite{langsch2019robotic}. To tackle the inter-patient variation, Virga~\emph{et al.} applied non-rigid registration to elastically register the patient to a generic MRI-based template using a Gaussian Mixture Model (GMM)~\cite{virga2016automatic}. Specifically for articulated joint motion, Jiang~\emph{et al.} proposed a non-rigid registration method to autonomously generate a scanning path for the limb examination based on a generic MRI template~\cite{jiang2022towards}. The robustness and effectiveness of this approach have been evaluated on different volunteers.

\par
The aforementioned approaches were proven to be robust in their scenarios, such as the aorta and limb artery examinations. However, for specific applications, such as thoracic examination or intervention within limited acoustic spaces, additional physical constraints should be further taken into consideration to develop an effective and robust system. Due to the high acoustic impedance of bone, an acoustic shadow will be cast and further hinder the visibility of the tissue underneath. To address this challenge, Jiang~\emph{et al.} first proposed a skeleton graph-based non-rigid registration method enabling the transferring of a pre-planned path in the limited intercostal space from CT template to patients~\cite{jiang2023skeleton}. To further address the cumbersome hyperparameters tuning, an evolved solution using a directed dense skeleton graph-based registration was proposed~\cite{jiang2023thoracic}. 

\par
Besides path planning, registration has also been seen as a potential and promising solution to address the complex issue of patient movement during the scanning process. To develop the motion-aware robotic sonographer, Jiang~\emph{et al.} presented a framework to autonomously detect a patient's motion and update the trajectory based on the registration between the markers' locations (sparse point cloud) obtained before and after the motion using an external camera~\cite{jiang2021motion}. Additionally, an advanced study using the dense point clouds extracted from an RGB-D camera was proposed, and validated on a human-like arm phantom with an uneven surface outliner~\cite{jiang2022precise}. 

\par
The aforementioned approaches are mainly developed based on the extracted point clouds. Such approaches are good for robotic applications by balancing the registration accuracy and computational efficiency. However, it usually cannot provide pixel-wise precise registration. To address this problem, Hennersperger~\emph{et al.} carried out an online US-MRI deformable registration by computing the $LC^2$ similarity to ensure the accurate path transferring~\cite{hennersperger2016towards}. The $LC^2$ was proposed for multi-modal registration, and it has been reported that it can provide robust and reliable descriptions of different modalities~\cite{wein2008automatic}. Besides, Modality-Independent Neighbourhood Descriptor (MIND) is another prominent metric developed on the same purpose~\cite{heinrich2012mind}. Most recently, Ronchetti~\emph{et al.} introduced a generic framework to create expressive cross-modal descriptors using CNN~\cite{ronchetti2023disa}. The results demonstrated that the proposed method can outperform the previous similarity method in terms of computational efficiency, and generalization capability across anatomies and imaging modalities.

\subsection{Direct Approach for Action Reasoning}
In this section, we discuss the recent advancements in robotic learning in RUSS. Different from the last section, the focus here is on works that deduce appropriate actions based on the interaction through methods such as reinforcement learning (RL) or learning from demonstrations (LfD).

\subsubsection{Learning from Interactions}
\par
RL is a famous interactive learning paradigm, which has achieved super-human performance in various games~\cite{mnih2015human}. The core idea of RL is to build a mapping between the observed states and the optimized actions through numerous interactions with the unknown environment, guided by predefined reward functions. This process can be modeled as a Markov decision process (MDP). 
With the development of deep learning, nowadays, neural networks are utilized as powerful approximators when facing complex state-action space. In the field of robotic US, initial attempts have been made to exploit the usage of RL. Unlike most RL tasks where the true state of the agent is directly observable, most US acquisition tasks can only be modeled as partially observable MDP. Thus, the success of the search is largely determined by whether the model can correctly infer the true state, i.e., the relative position between the probe and the scanning target, from its observations, i.e., US images. Hase~\emph{et al.} implemented a deep Q-learning model to navigate the US probe towards the sacrum of the spine, while only considering two translational movements~\cite{hase2020ultrasound}. Li~\emph{et al.} extended the dimension of the action space of the Q-learning method from two to five to complete a similar acquisition task for spine US~\cite{li2021autonomous}. Confidence map~\cite{karamalis2012ultrasound} was implemented as part of the reward function to include the image quality assessment in the optimization loop. Instead of using US images as the observation, Ning~\emph{et al.} utilized a scenery image as state representation and employed an RL-based proximal policy optimization approach to realize stable localization of the standard plane~\cite{ning2021autonomic}.
However, the generalization ability of the aforementioned approaches is limited by the insufficient number of training data. To address this problem, Bi~\emph{et al.} proposed to utilize a segmentation network to only keep the task-related features and wipe out the unnecessary background information from the US image~\cite{bi2022vesnet}. The training can then be performed with simulated artificial data rather than acquiring real patients' US volumes. Instead of using the segmented mask as an intermediate space to bridge between the simulation environment and real scenario, Li~\emph{et al.} proposed to use simulated US images from CT as an intermediate domain for training of RL agent~\cite{li2023style}. During the inference stage, the real US images are transferred to the US simulation domain using GAN so that the training of RL can be solely done in the CT space.

\subsubsection{Learning from Demonstrations}
\par
LfD is another approach to achieve autonomous US examinations, where the model learns directly from experts' demonstrations. LfD methods can be further categorized into guidance networks, where the model predicts actions based on US images, and assessment networks, which estimate preference scores for individual US images to reveal the intentions of sonographers during screening. As a valuable side benefit, the latter one can further be used to indicate the ``language of sonographer"~\cite{jiang2023intelligent}.

\par
In order to mimic how sonographers perform US examinations, Droste~\emph{et al.} proposed US-GuideNet to predict the action of the operator as well as the goal position of the standard plane using imitation learning~\cite{droste2020automatic}. The demonstrations consist of paired US frames and the orientations of the probe recorded from an IMU sensor. A gated recurrent unit was implemented to exploit the information of sequential data. However, it is still a big challenge to extract practical clues from the US images for action prediction. As a result, Men~\emph{et al.} integrated the gaze signal of the operator into the network to provide extra guidance~\cite{men2022multimodal}. The proposed Multimodal-GuideNet is able to simultaneously estimate the gaze of the operator and the probe motion. 
Instead of only providing instructions to the operator during scanning, Deng~\emph{et al.} implemented the learned policy from demonstrations directly to control the robot~\cite{deng2021learning}. The US frames together with contact forces and probe positions were recorded as expert demonstrations.

\par
Rather than cloning the movement of the expert from the given demonstrations, another way of achieving LfD is through inverse reinforcement learning (IRL)~\cite{abbeel2004apprenticeship}, which infers a reward function from demonstrations. Regarding US standard plane acquisition, such reward function refers to the preference of sonographers, demonstrating the anatomical and diagnostic value of the US image. The common assumption for IRL is the positive correlation between the reward value and the occurrence frequency of a given state in demonstrations~\cite{ziebart2008maximum}. However, in most cases, the US navigation is not optimal, which indicates the existence of a lot of redundant actions in the demonstrations. To address the problem of suboptimal demonstrations, Burke~\emph{et al.} proposed probabilistic temporal ranking method~\cite{burke2023learning}. It assumed that the US image in the latter stage of the demonstration is more likely to have higher diagnostic importance than the image in an earlier stage. Such an assumption is realistic for US screening, where the search normally terminates at the point when the standard plane is properly visualized. By relating the importance of US image with its relative position to the end point of the demonstration, Jiang~\emph{et al.} proposed global probabilistic spatial ranking~\cite{jiang2023intelligent}. A feature disentanglement technique was also introduced to the network to improve its performance on unseen scenarios of ex-vivo phantoms and in-vivo carotid US data.

\section{Advancements Towards Intelligence in Robotic Ultrasound}~\label{sec:IV_advanced_intelligence}


In this section, we demonstrate some important aspects that need to be properly addressed for RUSS, i.e., data scarcity, the necessity of integrating physics into the design of deep learning network, and efficient data representation.


\subsection{Data Scarcity}
For RUSS to be successfully applied in clinical practice, they must be both robust and reliable. However, achieving these attributes presents several challenges. A primary obstacle is the limited availability of ground truth data essential for training machine learning algorithms. Traditional approaches to data collection not only consume significant time and resources but often necessitate expert involvement for accurate labeling. This creates bottlenecks in developing machine learning models. This section presents methodologies to either circumvent or minimize the requirement for labeled data.

\subsubsection{Ultrasound Simulation}
By employing physics-based US simulations, researchers can generate a wealth of synthetic but realistic data, serving as ground truth for machine learning models. 
There are three main approaches for US simulation, i.e., interpolative methods, physics-based generative models, and deep learning-based techniques. These methods aim to generate realistic 2D or 3D images by capturing the complexities of tissue interactions with US waves.


\par
Interpolative methods, as implemented by Daulignac~\emph{et al.} \cite{d2006towards} and Goksel~\emph{et al.} \cite{goksel2009b}, serve as a fast and efficient means of image reconstruction. These techniques employ previously acquired US images and use them as a basis to generate new images through interpolation algorithms. The approach excels in quick image processing and realistic domain representation. However, it struggles to accurately portray directional artifacts, and its capability to produce novel views is constrained by the set of recorded images.

\par
When it comes to physics-based generative simulation, wave simulation serves as one of the most accurate, albeit computationally demanding, approaches. One of the most popular libraries, developed by Treeby~\emph{et al.}, is the k-wave toolbox \cite{treeby2010k}, which relies on solving coupled first-order acoustic equations in the k-space pseudo-spectral time domain. 

\par
Following a different, computationally less expensive approach, convolution-based methods excel in visualizing sub-wavelength tissue structures by convolving a Point Spread Function (PSF) with a scatterer map \cite{gao2009fast}. While effective for microscopic interactions, they lack accuracy in macroscopic correlations. Ray tracing, leveraging geometric optics principles, fills this gap by modeling US propagation as rays interacting with tissue. Combining convolution with ray tracing offers a comprehensive approach, capturing both microscopic and macroscopic interactions in US simulations~\cite{burger2012real}.



\par
Additionally, by leveraging pre-segmented medical imaging data, patient-specific US simulation can be acquired. For instance, Salehi~\emph{et al.} \cite{salehi2015patient} used segmented MRI data as input and applied convolutional ray tracing for appearance optimization. Velikova~\emph{et al.} used this approach for training a segmentation network \cite{velikova2022cactuss} only based on simulated US data, based on MRI, and associated labels. An enhanced version of the ray-tracing simulation is presented by Mattausch \emph{et al.} \cite{mattausch2018realistic}. By combining surface models with Monte Carlo surface interaction sampling for macroscopic interactions and convolution-based simulations for scattering phenomena an accurate representation of the underlying texture can be presented. 

\par
Deep learning-based techniques, particularly Generative Adversarial Networks (GANs), offer a data-driven mechanism for US image synthesis. Research such as that by Tom~\emph{et al.} \cite{tom2018simulating} and Hu~\emph{et al.} \cite{hu2017freehand} have exploited the capabilities of GANs to generate highly realistic US images. In addition, the application of CycleGANs, demonstrated by Vitale~\emph{et al.} \cite{vitale2020improving}, has shown potential in reconciling the discrepancies between simulated and real US images. This makes it a promising technique for training and validating machine learning models in robotic US applications.

\subsubsection{Data Augmentation}
In order to increase the diversity of training data, another intuitive way is data augmentation.
Out-of-domain and synthetic images are created based on the training data using either conventional image transformation methods or deep learning based generative models. As a traditional augmentation method, Zhang~\emph{et al.} proposed a deep stacked transformation framework, which has been tested on different medical imaging modalities and demonstrated promising results in increasing generalization capability~\cite{zhang2020generalizing}. 
Specialized in kidney US segmentation task, Yin~\emph{et al.} utilized a non-rigid image registration method to register and alter the shape of one kidney with other kidney samples in the training dataset so that the anatomical correctness of the augmented kidney can be guaranteed~\cite{yin2020automatic}. Lee~\emph{et al.} introduced a non-linear mixed-example augmentation method by fusing and mixing images from different classes to generate new training samples for standard plane classification of fetal US~\cite{lee2021principled}. 

\par
Fascinated by the superior performance of generative models in synthetic image generation, researchers have also implemented these models in the area of US image analysis~\cite{pang2021semi, tiago2022data}. 
In order to increase the interpretability during image generation, Shi~\emph{et al.} integrated expert annotations into GAN to generate anatomy- and domain-specific synthetic US images for thyroid nodule classification~\cite{shi2020knowledge}. 
For a similar application, Zhang~\emph{et al.} implemented a progressive GAN-based inpainting method to generate reliable structure and texture for thyroid US images analysis~\cite{zhang2021progressive}. 
Apart from GAN, variational auto-encoder (VAE) is another alternative for image synthesis. A conditional VAE was employed by Pesteie~\emph{et al.} to augment data for classification of spinal US, where the label information and the image embedding are disentangled to increase the manipulability during synthesis~\cite{pesteie2019adaptive}. In order to realize deformable augmentation for 3D US volumes, Wulff~\emph{et al.} implemented a conditional VAE to apply deformation on the 3D patch of US volumes~\cite{wulff2023towards}. 
However, the development and use of augmentation methods, which do not explicitly take the physics of US into account, could result in unrealistic and sometimes physically wrong US images. The addition of such images could even reduce the performance and robustness of the US image processing. By explicitly taking the particular physics of US into account, Tirindelli~\emph{et al.} introduced three physical-inspired augmentation strategies, i.e., deformation, reverberation, and signal-to-noise ratio, to generate plausible US images~\cite{tirindelli2021rethinking}.

\subsection{Physics-Inspired Machine Learning for US}\label{sec:physics-inspired}
In the previous sections, we have shown how the latest breakthrough in the field of machine learning and deep learning has elevated the conception of RUSS to a higher level.
Nonetheless, medical images are different from normal optical images. The intensity value of each pixel no longer represents the color only but also reflects the unique properties of human tissues. 
US, as a mechanical wave, has a much slower propagation speed compared to X-ray or light, which makes the resulting images very vulnerable to disturbances and full of noise. Such issues have posed extra challenges for a comprehensive understanding of US images. Thus, researchers have attempted to exploit extra clues from either physiological context or US formation principles. Physiology-inspired approaches refer to network architectures in which clinical prior knowledge is explicitly integrated to facilitate a deep understanding of the US images. 
On the other hand, taking the US formation principles into consideration when designing the network structures appears to be another promising solution. By doing so, the US images are no longer treated as a distribution of grayscale values but as a representation of human tissues.

\subsubsection{Physiology-Inspired Image Analysis}
Unlike normal optical images, US images are hard to interpret without sufficient medical knowledge. 
A deep understanding of anatomy is one of the key features that differentiates experienced sonographers from trainees. Inspired by the interpretation process of the medical experts, physiological knowledge is explicitly integrated into the deep learning based US analysis network.
In order to improve the segmentation performance of vessels in US images, Jiang~\emph{et al.} proposed OF-UNet, which explicitly utilized the continuity of vessels~\cite{jiang2022towards}. 
Guidance is provided to the network based on the optical flow image and the predicted segmentation mask from the previous frame. During breast US examination, sonographers utilize the information of previous frames to determine whether a lesion-like anatomy is indeed a diseased tissue or not. Based on such observations, Yu~\emph{et al.} proposed UltraDet, in which an inverse optical flow progress was implemented to trace back in previous frames for detecting negative symptoms, that indicate false-positive classifications~\cite{yu2023mining}.
To produce anatomically plausible semantic segmentation of cardiac US, Painchaud~\emph{et al.} proposed a framework to guarantee the anatomical correctness of resulting segmentation maps~\cite{painchaud2020cardiac}.
By transferring all the ground truth labels from image space to latent space using a conditioned VAE, a manifold of valid cardiac shapes in the feature space can be determined. Then, a mapping between the anatomically incorrect segmentation mask and the manifold of valid shapes is performed to improve the shape correctness.
To correct the force-induced US image deformation, Jiang~\emph{et al.} explicitly considered tissue stiffness maps to compute an accurate and consistent anatomic geometry in RUSS~\cite{jiang2023defcor, jiang2021deformation}.

\par
To resolve the challenge of detecting tiny tissues in US images, e.g., nerves, Dou~\emph{et al.} proposed a pipeline, which utilized the anatomical knowledge to provide region proposals~\cite{dou2022localizing}. Guided by the segmented surrounding organs, i.e., carotid artery, thyroid, and trachea, a Bayesian shape alignment module is applied to propose the region proposal for locating recurrent laryngeal nerve.
In order to generate temporal consistent segmentation results for cardiac US, Painchaud~\emph{et al.} proposed a framework to perform post-processing on the segmentation results~\cite{painchaud2022echocardiography}. An autoencoder was introduced to map the segmentation masks into an interpretable latent space. For a sequence of cardiac US images, a smoothing process is applied to each attribute of the latent representations until a temporal consistency metric is satisfied. The periodic motions of arteries are frequently utilized by doctors as a strong prior for detection and localization in clinical practice. Huang~\emph{et al.} employed such prior knowledge into the network design to facilitate the segmentation results of small arteries, i.e., radial artery~\cite{huang2023motion}. Motion magnification techniques were introduced to extract a pulsation map, which highlights the region where the periodic motions occur. The pulsation map then served as an attention map to guide the segmentation network.

\subsubsection{Ultrasound Physics-Inspired Image Analysis}
The gray scale values in US images represent the intensity of the reflected US waves. For visualization reasons, the source signals are mapped into a color space between 0 and 255, so that they are interpretable by humans. A lot of information is lost during such a visualization process. However, this is not necessarily true for machines. With growing computational power, it is becoming feasible to learn how to interpret the source data directly. Therefore, Gare~\emph{et al.} proposed W-Net, combining raw radio frequency data with B-mode image to optimize the segmentation results~\cite{gare2022w}. By extracting information from the radio frequency data, W-Net is able to surpass the performance of a U-Net with a smaller network size. 
Another distinct property inherent to US is the Doppler effect. Doppler images are widely used in clinical practice to detect and measure blood flow. It has also been applied to provide weak supervision for the vessel segmentation network~\cite{ning2023doppler}.
Jiang~\emph{et al.} later combined Doppler images with the B-mode US images to provide extra guidance to the network for radial artery segmentation~\cite{jiang2023dopus}. Since the Doppler effect is unstable during autonomous US scanning, a RUSS was designed with the capability to reacquire a high-quality Doppler signal by performing out-of-plane fan motion.

\par
A significant challenge in US imaging is the scarcity or absence of information in regions with attenuation or shadowing, a factor that substantially impedes the interpretability of these areas. Based on the principles of US wave propagation, Karamalis~\emph{et al.} introduced confidence maps, which modeled the confidence assessment task as a random walk problem~\cite{karamalis2012ultrasound}. The resulting confidence map was then utilized by several RUSS as a criterion for US image quality optimization~\cite{jiang2022precise, li2021image, chatelain2017confidence}.
Later, Klein~\emph{et al.} proposed to integrate radio frequency data into the random walk-based confidence estimation~\cite{klein2015rf}.
Meng~\emph{et al.} introduced a learning-based method to estimate the shadow areas for fetal US using a weak-supervised learning approach since it is not a feasible practice to provide pixel-level shadow annotations from human operators~\cite{meng2019weakly}.

\par
Enlightened by physics-based US simulation from CT segmentation maps, Velikova~\emph{et al.} proposed to compose an intermediate CT-US space through US simulation parameterization~\cite{velikova2022cactuss}. The segmentation is performed solely in the intermediate space, while CT and US images are transferred to such common anatomical space through a specific parameterized US simulator during training and a GAN during inference, respectively. One step further, instead of using hand-crafted parameterization for CT to US simulation, Velikova~\emph{et al.} introduced a fully differentiable US simulator and realized an end-to-end training setting~\cite{velikova2023lotus}. The network is then able to optimize the intermediate space for different anatomy separately.
Inspired by the neural radiance field~\cite{mildenhall2021nerf}, Wysocki~\emph{et al.} presented Ultra-NeRF, which learned an intrinsic representation of each point in the US volume based on US physics, so that US images can be generated from new viewpoints~\cite{wysocki2023ultra}. 

\subsection{Efficient Data Representation}
The effectiveness of a CNN is largely dependent on its ability to extract distinct features. The redundancy in feature space largely hinders the inference performance of deep learning models, especially when the diversity of the training dataset is limited.
In order to increase the conciseness of the data representation, a disentanglement between task-related and task-irrelevant features is an essential step towards robust and reliable intelligent assessment of US images.
Using V-Net~\cite{milletari2016v} as the backbone, Degel~\emph{et al.} proposed a 3D segmentation framework, where a discriminator is implemented to predict the specific domain of the input US volume and the corresponding adversarial loss forces the segmentation network to ignore the domain information~\cite{degel2018domain}. 
Meng~\emph{et al.} also utilized adversarial loss to regularize the network to extract distinct features for corresponding classification tasks separately~\cite{meng2019representation}. The performance of the proposed segmentation network was validated on fetal US standard plane classification and shadow detection tasks. 
To increase the cross-domain robustness of the network, Meng~\emph{et al.} implemented feature space clustering and demonstrated promising results in handling the domain shift in fetal US standard planes detection~\cite{meng2020unsupervised}.
Combining feature space alignment and adversarial learning strategy, Ying~\emph{et al.} proposed a classification scheme for benign and malignant thyroid nodules and showed high generalization ability across different US machines~\cite{ying2022multi}.
For the same task, instead of using a distance-based feature alignment method, Zhang~\emph{et al.} used a graph convolution network~\cite{morris2019weisfeiler} to assess the relations between the source and target domain data and further realize cross-device generalization~\cite{zhang2023semi}.
Another more intuitive way to decouple the domain- and task-related features is through minimizing a metric that can directly measure the shared information between the two variables. To this end, mutual information~\cite{belghazi2018mutual}, which measures the dependency between two random variables, has been applied in various classification tasks in computer vision area~\cite{cha2022domain, liu2021mutual, peng2019domain}. In the US field, Meng~\emph{et al.} realized unseen domain generalization for fetal US standard plane detection through mutual information minimization~\cite{meng2020mutual}. Following a similar concept, Bi~\emph{et al.} extended the usage of mutual information to segmentation networks by introducing a cross-reconstruction strategy~\cite{bi2023mi}.

\section{Open Challenges and Future Perspectives}~\label{Sec:V_future_studies}
\par
\revision{Significant progress has been made toward achieving the ultimate goal of developing intelligent robotic sonographers by enhancing the understanding of the environment using advanced machine learning algorithms. However, there are still some challenges remaining in the research community, such as ethical and legal issues.
Apart from that, emerging hardware design in the US area has revealed brand-new potential for future research.
In this section, we will discuss the ethical problems and provide insights for potential directions to stimulate future studies.
}

\subsection{Ethics and Regulations}
\par
Given the rising enthusiasm within the market, scientific community, and the medical field, ethical and regulatory guidance is becoming more and more important. A clear responsibility definition of such emerging systems could boost the research in the field of clinical translation, in order to bring the prototypes from the laboratories to the market. The demand for proper regulations has been discussed and emphasized by pioneers in the area of applied medical robotics~\cite{fichtinger2022image, khamis2019ai, haidegger2019autonomy, prestes2021first, haidegger2022robot, jiang2023robotic}. 

\par
To tackle the ethical, regulatory, and legal issues for autonomous medical robots, Yang~\emph{et al.} defined six levels of autonomy: no autonomy, robot assistance, task autonomy,
conditional autonomy, high autonomy, and full autonomy~\cite{yang2017medical}. This taxonomy has been widely used to categorize the relevant techniques to their respective level~\cite{attanasio2021autonomy, haidegger2019autonomy}. Different degrees of autonomy require distinct regulatory frameworks. In the case of medical robots with low autonomy levels, accountability primarily rests with the clinicians who maintain complete control over the system. Due to this well-defined attribution of responsibility, the transition to commercialization for such systems tends to be more successful compared to others. The da Vinci robot serves as the most prominent example of this success.


\par
Concerning the regulation of medical systems that exhibit a certain level of autonomy, two leading Standard Development Organizations — the International Organization for Standardization (ISO) and the International Electrotechnical Commission (IEC) — have published their inaugural joint standard (IEC/TR 60601-4-1) to address the safety, transparency, and trustworthiness of contemporary medical devices~\cite{haidegger2019autonomy}. Unlike surgical robots, robotic sonographers are non-invasive and interact with patients who are not under anesthesia; thus, more specialized regulations are necessary. 



\subsection{Emerging Ultrasound Imaging System}
\par
\revision{
The fundamental developments in new types of US sensing hardware will significantly impact current clinical intervention procedures and bring new opportunities. In this section, we highlight the emerging optical US systems and soft ultrasonic patches.
}

\subsubsection{Optical Ultrasound}
To enhance the versatility of US technologies, integrating optical systems with acoustic methods offers an intriguing path forward. By leveraging laser-induced thermal expansion within biological tissues, optoacoustic imaging enables high-resolution, molecularly-specific imaging. This capability is crucial for distinguishing various tissue types, identifying pathological conditions, and pinpointing specific molecular markers \cite{karlas2021optoacoustic}. 
Building on recent progress, advancements in all-optical US (OpUS) transducers show particular promise for use in minimally invasive surgery. OpUS generates US via a fiber optic transducer using the photoacoustic effect, and captures tissue reflections through optical interferometry, akin to OCT methods \cite{little2020optical}. These transducers excel in their remarkable miniaturization while preserving high-resolution imaging and mechanical flexibility, making them especially useful in endovascular applications like thrombus imaging \cite{zhang2023miniaturised}.

Notably, this technology allows real-time feedback, creating opportunities for robotic monitoring in surgical procedures and image-based diagnostics. A research team at Johns Hopkins University was the first to demonstrate a robotic tracking system guided by optoacoustic signals. The approach involves embedding an optical fiber into the distal end of a surgical instrument. This setup allows a robotic manipulator, equipped with a US probe, to monitor and trace the position of the tool tip through analysis of the optoacoustic signal. Such innovation has found applications in robotic catheter \cite{graham2019vivo} and needle tracking \cite{shubert2017photoacoustic}.

\subsubsection{Ultrasonic Patch}
Given the recent advancements in unconventional US sensors such as US patches~\cite{hu2018stretchable, wang2021continuous}, the field of US applications could be significantly expanded. These emerging soft US patch technologies, thanks to their size and flexibility, facilitate extended monitoring of patients. This allows for continuous capture of physiological data, aiding in accurate clinical diagnoses and close tracking of disease progression. Such extended monitoring was previously expensive and inefficient when utilizing traditional, bulky US probes.

\par
In 2018, Hu~\emph{et al} presented a soft US patch consisting of $10\times 10$ array of piezoelectric transducers for 3D imaging of deep tissues~\cite{hu2018stretchable}. They reported that the patch holds excellent electromechanical coupling, minimal cross-talk, and more than 50\% stretchability. Wang~\emph{et al.} applied a skin-conformal US patch array to monitor hemodynamic signals from tissues up to $14~cm$ beneath the skin~\cite{wang2021continuous}. In another study, Wang~\emph{et al.} present a bioadhesive US array to provide $48$ hours of continuous imaging of various organs, including blood vessels, muscle, heart, diaphragm, lung, etc.~\cite{wang2022bioadhesive}. Focusing on cardiac imaging, Hu~\emph{et al.} proposed a wearable US patch imager which is used to extract continuous signals for evaluating cardiac functions~\cite{hu2023wearable}. 
When paired with a deep learning model for the automated segmentation of ventricular volume, this setup enables real-time calculations of essential cardiac performance metrics, including stroke volume, cardiac output, and ejection fraction.
To provide wireless communication, Lin~\emph{et al.} designed a miniaturized flexible control circuit which directly interfaces with the US patch~\cite{lin2023fully}. This allows continuous imaging of moving objects. These fundamental advancements in US sensors could unlock numerous opportunities for revolutionizing the techniques of robot-assisted US imaging. 

\section{Conclusion}
\par
This review has focused on highlighting the significance of machine learning and deep learning technologies in enhancing the ``intelligence" of RUSS. To provide an overview of the ongoing developments in the field, we initially discussed the specific robotic mechanisms and control systems employed in RUSS, as well as their clinical applications. Modern robotic systems and control methods have already demonstrated superhuman levels of reproducibility. However, they still face challenges in understanding the physiological aspects of human tissue and diseases, different imaging modalities, imaging physics, and artifacts. Additionally, there is a need to improve their ability to promptly reason and make the correct decisions in response to expected or unexpected changes during acquisition. To achieve this, we discussed two categories of methods for action reasoning, one based on implicitly interpreting input data and the other on explicitly interpreting it. Towards the conclusion of the paper, we also discussed the open challenges and future perspectives. The development and deployment of intelligent RUSS will be a challenging and exciting endeavor. We believe robotic sonography and autonomous US examination represent the next frontier in medical robotics and hold significant promise in future clinical practices. Furthermore, we highlighted the crucial importance of advancements in the area of ethics and regulations and also emphasized the need for collaborative efforts from the scientific, industrial, and clinical communities.
\section*{DISCLOSURE STATEMENT}
The authors are not aware of any affiliations, memberships, funding, or financial holdings that
might be perceived as affecting the objectivity of this review. 

\bibliographystyle{ar-style3.bst}
\bibliography{references}

\begin{thebibliography}{149}
\expandafter\ifx\csname natexlab\endcsname\relax\def\natexlab#1{#1}\fi

\bibitem{dupont2021decade}
Dupont PE, Nelson BJ, Goldfarb M, Hannaford B, Menciassi A, et~al. 2021.
A decade retrospective of medical robotics research from 2010 to 2020.
\textit{Science robotics} 6(60):eabi8017

\bibitem{yang2017medical}
Yang GZ, Cambias J, Cleary K, Daimler E, Drake J, et~al. 2017.
Medical robotics—regulatory, ethical, and legal considerations for increasing levels of autonomy

\bibitem{yang2020combating}
Yang GZ, J.~Nelson B, Murphy RR, Choset H, Christensen H, et~al. 2020.
Combating covid-19—the role of robotics in managing public health and infectious diseases

\bibitem{yip2023artificial}
Yip M, Salcudean S, Goldberg K, Althoefer K, Menciassi A, et~al. 2023.
Artificial intelligence meets medical robotics.
\textit{Science} 381(6654):141--146

\bibitem{zemmar2020rise}
Zemmar A, Lozano AM, Nelson BJ. 2020.
The rise of robots in surgical environments during covid-19.
\textit{Nature Machine Intelligence} 2(10):566--572

\bibitem{jiang2023robotic}
Jiang Z, Salcudean SE, Navab N. 2023.
Robotic ultrasound imaging: State-of-the-art and future perspectives.
\textit{Medical Image Analysis} :102878

\bibitem{von2021medical}
von Haxthausen F, B{\"o}ttger S, Wulff D, Hagenah J, Garc{\'\i}a-V{\'a}zquez V, Ipsen S. 2021.
Medical robotics for ultrasound imaging: current systems and future trends.
\textit{Current robotics reports} 2:55--71

\bibitem{li2021overview}
Li K, Xu Y, Meng MQH. 2021.
An overview of systems and techniques for autonomous robotic ultrasound acquisitions.
\textit{IEEE Transactions on Medical Robotics and Bionics} 3(2):510--524

\bibitem{navab2016personalized}
Navab N, Hennersperger C, Frisch B, F{\"u}rst B. 2016.
Personalized, relevance-based multimodal robotic imaging and augmented reality for computer assisted interventions

\bibitem{salcudean1999robot}
Salcudean SE, Bell G, Bachmann S, Zhu WH, Abolmaesumi P, Lawrence PD. 1999.
\textit{Robot-assisted diagnostic ultrasound--design and feasibility experiments}.
In \textit{Medical Image Computing and Computer-Assisted Intervention--MICCAI’99: Second International Conference, Cambridge, UK, September 19-22, 1999. Proceedings 2}, pp.  1062--1071. Springer

\bibitem{jiang2020automatic}
Jiang Z, Grimm M, Zhou M, Esteban J, Simson W, et~al. 2020{\natexlab{a}}.
Automatic normal positioning of robotic ultrasound probe based only on confidence map optimization and force measurement.
\textit{IEEE Robotics and Automation Letters} 5(2):1342--1349

\bibitem{jiang2020automatic_TIE}
Jiang Z, Grimm M, Zhou M, Hu Y, Esteban J, Navab N. 2020{\natexlab{b}}.
Automatic force-based probe positioning for precise robotic ultrasound acquisition.
\textit{IEEE Transactions on Industrial Electronics} 68(11):11200--11211

\bibitem{zhang2023robotic}
Zhang T, Pang Y, Zeng T, Wang G, Yin S, et~al. 2023{\natexlab{a}}.
Robotic drilling for the chinese chang’e 5 lunar sample-return mission.
\textit{The International Journal of Robotics Research} 42(8):586--613

\bibitem{li2021self}
Li G, Chen X, Zhou F, Liang Y, Xiao Y, et~al. 2021{\natexlab{a}}.
Self-powered soft robot in the mariana trench.
\textit{Nature} 591(7848):66--71

\bibitem{jiang2023intelligent}
Jiang Z, Bi Y, Zhou M, Hu Y, Burke M, Navab N. 2023{\natexlab{a}}.
Intelligent robotic sonographer: Mutual information-based disentangled reward learning from few demonstrations.
\textit{arXiv preprint arXiv:2307.03705}

\bibitem{baumgartner2017sononet}
Baumgartner CF, Kamnitsas K, Matthew J, Fletcher TP, Smith S, et~al. 2017.
Sononet: real-time detection and localisation of fetal standard scan planes in freehand ultrasound.
\textit{IEEE transactions on medical imaging} 36(11):2204--2215

\bibitem{droste2020automatic}
Droste R, Drukker L, Papageorghiou AT, Noble JA. 2020.
\textit{Automatic probe movement guidance for freehand obstetric ultrasound}.
In \textit{International Conference on Medical Image Computing and Computer-Assisted Intervention}, pp.  583--592. Springer

\bibitem{mustafa2013development}
Mustafa ASB, Ishii T, Matsunaga Y, Nakadate R, Ishii H, et~al. 2013.
\textit{Development of robotic system for autonomous liver screening using ultrasound scanning device}.
In \textit{2013 IEEE international conference on robotics and biomimetics (ROBIO)}, pp.  804--809. IEEE

\bibitem{giuliani2020user}
Giuliani M, Szcz{\k{e}}{\'s}niak-Sta{\'n}czyk D, Mirnig N, Stollnberger G, Szyszko M, et~al. 2020.
User-centred design and evaluation of a tele-operated echocardiography robot.
\textit{Health and Technology} 10:649--665

\bibitem{ma2021autonomous}
Ma X, Zhang Z, Zhang HK. 2021.
\textit{Autonomous scanning target localization for robotic lung ultrasound imaging}.
In \textit{2021 IEEE/RSJ International Conference on Intelligent Robots and Systems (IROS)}, pp.  9467--9474. IEEE

\bibitem{tan2022flexible}
Tan J, Li B, Li Y, Li B, Chen X, et~al. 2022.
A flexible and fully autonomous breast ultrasound scanning system.
\textit{IEEE Transactions on Automation Science and Engineering}

\bibitem{huang2023motion}
Huang D, Bi Y, Navab N, Jiang Z. 2023.
Motion magnification in robotic sonography: Enabling pulsation-aware artery segmentation.
\textit{IROS 2023}

\bibitem{zielke2022rsv}
Zielke J, Eilers C, Busam B, Weber W, Navab N, Wendler T. 2022.
Rsv: Robotic sonography for thyroid volumetry.
\textit{IEEE Robotics and Automation Letters} 7(2):3342--3348

\bibitem{esmaeeli2020robotically}
Esmaeeli S, Hrdlicka CM, Bastos AB, Wang J, Gomez-Paz S, et~al. 2020.
Robotically assisted transcranial doppler with artificial intelligence for assessment of cerebral vasospasm after subarachnoid hemorrhage.
\textit{Journal of Neurocritical Care} 13(1):32--40

\bibitem{tirindelli2020force}
Tirindelli M, Victorova M, Esteban J, Kim ST, Navarro-Alarcon D, et~al. 2020.
Force-ultrasound fusion: Bringing spine robotic-us to the next “level”.
\textit{IEEE Robotics and Automation Letters} 5(4):5661--5668

\bibitem{virga2016automatic}
Virga S, Zettinig O, Esposito M, Pfister K, Frisch B, et~al. 2016.
\textit{Automatic force-compliant robotic ultrasound screening of abdominal aortic aneurysms}.
In \textit{2016 IEEE/RSJ International Conference on Intelligent Robots and Systems (IROS)}, pp.  508--513. IEEE

\bibitem{shida2021heart}
Shida Y, Tsumura R, Watanabe T, Iwata H. 2021.
\textit{Heart Position Estimation based on Bone Distribution toward Autonomous Robotic Fetal Ultrasonography}.
In \textit{2021 IEEE International Conference on Robotics and Automation (ICRA)}, pp.  11393--11399. IEEE

\bibitem{stilli2019pneumatically}
Stilli A, Dimitrakakis E, D'Ettorre C, Tran M, Stoyanov D. 2019.
Pneumatically attachable flexible rails for track-guided ultrasound scanning in robotic-assisted partial nephrectomy—a preliminary design study.
\textit{IEEE Robotics and Automation Letters} 4(2):1208--1215

\bibitem{hungr20123}
Hungr N, Baumann M, Long JA, Troccaz J. 2012.
A 3-d ultrasound robotic prostate brachytherapy system with prostate motion tracking.
\textit{IEEE Transactions on Robotics} 28(6):1382--1397

\bibitem{jiang2022towards}
Jiang Z, Gao Y, Xie L, Navab N. 2022{\natexlab{a}}.
Towards autonomous atlas-based ultrasound acquisitions in presence of articulated motion.
\textit{IEEE Robotics and Automation Letters} 7(3):7423--7430

\bibitem{li2022dual}
Li T, Meng X, Tavakoli M. 2022.
Dual mode phri-telehri control system with a hybrid admittance-force controller for ultrasound imaging.
\textit{Sensors} 22(11):4025

\bibitem{chatelain2017confidence}
Chatelain P, Krupa A, Navab N. 2017.
Confidence-driven control of an ultrasound probe.
\textit{IEEE Transactions on Robotics} 33(6):1410--1424

\bibitem{tsumura2020robotic}
Tsumura R, Iwata H. 2020.
Robotic fetal ultrasonography platform with a passive scan mechanism.
\textit{International Journal of Computer Assisted Radiology and Surgery} 15:1323--1333

\bibitem{wang2019analysis}
Wang S, Housden RJ, Noh Y, Singh A, Lindenroth L, et~al. 2019.
Analysis of a customized clutch joint designed for the safety management of an ultrasound robot.
\textit{Applied Sciences} 9(9):1900

\bibitem{welleweerd2020automated}
Welleweerd MK, de~Groot AG, de~Looijer S, Siepel FJ, Stramigioli S. 2020.
\textit{Automated robotic breast ultrasound acquisition using ultrasound feedback}.
In \textit{2020 IEEE international conference on robotics and automation (ICRA)}, pp.  9946--9952. IEEE

\bibitem{facundo2020design}
Facundo-Flores L, Treesatayapun C, Baltazar A. 2020.
Design of a pose and force controller for a robotized ultrasonic probe based on neural networks and stochastic gradient approximation.
\textit{IEEE Sensors Journal} 21(5):6224--6233

\bibitem{wang2023compliant}
Wang Y, Liu T, Hu X, Yang K, Zhu Y, Jin H. 2023{\natexlab{a}}.
Compliant joint based robotic ultrasound scanning system for imaging human spine.
\textit{IEEE Robotics and Automation Letters}

\bibitem{bao2023novel}
Bao X, Wang S, Zheng L, Housden RJ, Hajnal JV, Rhode K. 2023.
A novel ultrasound robot with force/torque measurement and control for safe and efficient scanning.
\textit{IEEE transactions on instrumentation and measurement} 72:1--12

\bibitem{goel2022autonomous}
Goel R, Abhimanyu F, Patel K, Galeotti J, Choset H. 2022.
\textit{Autonomous ultrasound scanning using bayesian optimization and hybrid force control}.
In \textit{2022 International Conference on Robotics and Automation (ICRA)}, pp.  8396--8402. IEEE

\bibitem{napoli2018hybrid}
Napoli ME, Freitas C, Goswami S, McAleavey S, Doyley M, Howard TM. 2018.
\textit{Hybrid Force/Velocity Control “With Compliance Estimation via Strain Elastography for Robot Assisted Ultrasound Screening}.
In \textit{2018 7th IEEE International Conference on Biomedical Robotics and Biomechatronics (Biorob)}, pp.  1266--1273. IEEE

\bibitem{dyck2022impedance}
Dyck M, Sachtler A, Klodmann J, Albu-Sch{\"a}ffer A. 2022.
Impedance control on arbitrary surfaces for ultrasound scanning using discrete differential geometry.
\textit{IEEE Robotics and Automation Letters} 7(3):7738--7746

\bibitem{fang2017force}
Fang TY, Zhang HK, Finocchi R, Taylor RH, Boctor EM. 2017.
Force-assisted ultrasound imaging system through dual force sensing and admittance robot control.
\textit{International journal of computer assisted radiology and surgery} 12:983--991

\bibitem{wang2023task}
Wang J, Lu C, Lv Y, Yang S, Zhang M, Shen Y. 2023{\natexlab{b}}.
Task space compliant control and six-dimensional force regulation toward automated robotic ultrasound imaging.
\textit{IEEE Transactions on Automation Science and Engineering}

\bibitem{guerrero2003deep}
Guerrero J, Salcudean S, McEwen JA, Masri BA, Nicolaou S. 2003.
\textit{Deep venous thrombosis screening system using numerical measures}.
In \textit{Proceedings of the 25th Annual International Conference of the IEEE Engineering in Medicine and Biology Society (IEEE Cat. No. 03CH37439)}, vol.~1, pp.  894--897. IEEE

\bibitem{duan2022ultrasound}
Duan A, Victorova M, Zhao J, Sun Y, Zheng Y, Navarro-Alarcon D. 2022.
Ultrasound-guided assistive robots for scoliosis assessment with optimization-based control and variable impedance.
\textit{IEEE Robotics and Automation Letters} 7(3):8106--8113

\bibitem{xiao2021visual}
Xiao S, Wang C, Shi Y, Yu J, Xiong L, et~al. 2021.
\textit{Visual optimization of ultrasound-guided robot-assisted procedures using variable impedance control}.
In \textit{2021 WRC Symposium on Advanced Robotics and Automation (WRC SARA)}, pp.  128--133. IEEE

\bibitem{santos2018computed}
Santos L, Cortes{\~a}o R. 2018.
Computed-torque control for robotic-assisted tele-echography based on perceived stiffness estimation.
\textit{IEEE Transactions on Automation Science and Engineering} 15(3):1337--1354

\bibitem{kim2019gain}
Kim YJ, Park CK, Kim KG. 2019.
Gain determination of feedback force for an ultrasound scanning robot using genetic algorithm.
\textit{International Journal of Computer Assisted Radiology and Surgery} 14:797--807

\bibitem{abbas2021event}
Abbas M, Al~Issa S, Dwivedy SK. 2021.
Event-triggered adaptive hybrid position-force control for robot-assisted ultrasonic examination system.
\textit{Journal of Intelligent \& Robotic Systems} 102:1--19

\bibitem{wang2022full}
Wang Z, Zhao B, Zhang P, Yao L, Wang Q, et~al. 2022{\natexlab{a}}.
Full-coverage path planning and stable interaction control for automated robotic breast ultrasound scanning.
\textit{IEEE Transactions on Industrial Electronics} 70(7):7051--7061

\bibitem{raina2023deep}
Raina D, Chandrashekhara S, Voyles R, Wachs J, Saha SK. 2023.
\textit{Deep Kernel and Image Quality Estimators for Optimizing Robotic Ultrasound Controller using Bayesian Optimization}.
In \textit{2023 International Symposium on Medical Robotics (ISMR)}, pp.  1--7. IEEE

\bibitem{akbari2021robotic}
Akbari M, Carriere J, Meyer T, Sloboda R, Husain S, et~al. 2021{\natexlab{a}}.
Robotic ultrasound scanning with real-time image-based force adjustment: quick response for enabling physical distancing during the covid-19 pandemic.
\textit{Frontiers in Robotics and AI} 8:645424

\bibitem{sutedjo2022acoustic}
Sutedjo V, Tirindelli M, Eilers C, Simson W, Busam B, Navab N. 2022.
Acoustic shadowing aware robotic ultrasound: Lighting up the dark.
\textit{IEEE Robotics and Automation Letters} 7(2):1808--1815

\bibitem{merouche2015robotic}
Merouche S, Allard L, Montagnon E, Soulez G, Bigras P, Cloutier G. 2015.
A robotic ultrasound scanner for automatic vessel tracking and three-dimensional reconstruction of b-mode images.
\textit{IEEE transactions on ultrasonics, ferroelectrics, and frequency control} 63(1):35--46

\bibitem{akbari2021robot}
Akbari M, Carriere J, Sloboda R, Meyer T, Usmani N, et~al. 2021{\natexlab{b}}.
\textit{Robot-assisted Breast Ultrasound Scanning Using Geometrical Analysis of the Seroma and Image Segmentation}.
In \textit{2021 IEEE/RSJ International Conference on Intelligent Robots and Systems (IROS)}, pp.  3784--3791. IEEE

\bibitem{liu2019deep}
Liu S, Wang Y, Yang X, Lei B, Liu L, et~al. 2019.
Deep learning in medical ultrasound analysis: a review.
\textit{Engineering} 5(2):261--275

\bibitem{ronneberger2015u}
Ronneberger O, Fischer P, Brox T. 2015.
\textit{U-net: Convolutional networks for biomedical image segmentation}.
In \textit{MICCAI}, pp.  234--241. Springer

\bibitem{jiang2021autonomous}
Jiang Z, Li Z, Grimm M, Zhou M, Esposito M, et~al. 2021{\natexlab{a}}.
Autonomous robotic screening of tubular structures based only on real-time ultrasound imaging feedback.
\textit{IEEE Transactions on Industrial Electronics} 69(7):7064--7075

\bibitem{koskinopoulou2023dual}
Koskinopoulou M, Acemoglu A, Penza V, Mattos LS. 2023.
\textit{Dual Robot Collaborative System for Autonomous Venous Access Based on Ultrasound and Bioimpedance Sensing Technology}.
In \textit{2023 IEEE International Conference on Robotics and Automation (ICRA)}, pp.  4648--4653. IEEE

\bibitem{chen2023fully}
Chen M, Huang Y, Chen J, Zhou T, Chen J, Liu H. 2023.
\textit{Fully Robotized 3D Ultrasound Image Acquisition for Artery}.
In \textit{2023 IEEE International Conference on Robotics and Automation (ICRA)}, pp.  2690--2696. IEEE

\bibitem{chen2021semi}
Chen Y, Wang Y, Lai B, Chen Z, Cao X, et~al. 2021.
\textit{Semi-supervised vein segmentation of ultrasound images for autonomous venipuncture}.
In \textit{2021 IEEE/RSJ International Conference on Intelligent Robots and Systems (IROS)}, pp.  9475--9481. IEEE

\bibitem{jiang2023dopus}
Jiang Z, Duelmer F, Navab N. 2023.
Dopus-net: Quality-aware robotic ultrasound imaging based on doppler signal.
\textit{IEEE Transactions on Automation Science and Engineering}

\bibitem{ning2023autonomous}
Ning G, Liang H, Zhang X, Liao H. 2023{\natexlab{a}}.
Autonomous robotic ultrasound vascular imaging system with decoupled control strategy for external-vision-free environments.
\textit{IEEE Transactions on Biomedical Engineering}

\bibitem{velikova2022cactuss}
Velikova Y, Simson W, Salehi M, Azampour MF, Paprottka P, Navab N. 2022.
\textit{CACTUSS: Common Anatomical CT-US Space for US Examinations}.
In \textit{MICCAI}, pp.  492--501. Springer

\bibitem{velikova2023lotus}
Velikova Y, Azampour MF, Simson W, Duque VG, Navab N. 2023.
Lotus: Learning to optimize task-based us representations.
\textit{MICCAI 2023}

\bibitem{che2017ultrasound}
Che C, Mathai TS, Galeotti J. 2017.
Ultrasound registration: A review.
\textit{Methods} 115:128--143

\bibitem{hennersperger2016towards}
Hennersperger C, Fuerst B, Virga S, Zettinig O, Frisch B, et~al. 2016.
Towards mri-based autonomous robotic us acquisitions: a first feasibility study.
\textit{IEEE transactions on medical imaging} 36(2):538--548

\bibitem{langsch2019robotic}
Langsch F, Virga S, Esteban J, G{\"o}bl R, Navab N. 2019.
\textit{Robotic ultrasound for catheter navigation in endovascular procedures}.
In \textit{2019 IEEE/RSJ International Conference on Intelligent Robots and Systems (IROS)}, pp.  5404--5410. IEEE

\bibitem{jiang2023skeleton}
Jiang Z, Li X, Zhang C, Bi Y, Stechele W, Navab N. 2023{\natexlab{b}}.
Skeleton graph-based ultrasound-ct non-rigid registration.
\textit{IEEE Robotics and Automation Letters}

\bibitem{jiang2023thoracic}
Jiang Z, Li C, Li X, Navab N. 2023{\natexlab{c}}.
Thoracic cartilage ultrasound-ct registration using dense skeleton graph.
\textit{IROS 2023}

\bibitem{jiang2021motion}
Jiang Z, Wang H, Li Z, Grimm M, Zhou M, et~al. 2021{\natexlab{b}}.
\textit{Motion-aware robotic 3D ultrasound}.
In \textit{2021 IEEE International Conference on Robotics and Automation (ICRA)}, pp.  12494--12500. IEEE

\bibitem{jiang2022precise}
Jiang Z, Danis N, Bi Y, Zhou M, Kroenke M, et~al. 2022{\natexlab{b}}.
Precise repositioning of robotic ultrasound: Improving registration-based motion compensation using ultrasound confidence optimization.
\textit{IEEE Transactions on Instrumentation and Measurement} 71:1--11

\bibitem{wein2008automatic}
Wein W, Brunke S, Khamene A, Callstrom MR, Navab N. 2008.
Automatic ct-ultrasound registration for diagnostic imaging and image-guided intervention.
\textit{Medical image analysis} 12(5):577--585

\bibitem{heinrich2012mind}
Heinrich MP, Jenkinson M, Bhushan M, Matin T, Gleeson FV, et~al. 2012.
Mind: Modality independent neighbourhood descriptor for multi-modal deformable registration.
\textit{Medical image analysis} 16(7):1423--1435

\bibitem{ronchetti2023disa}
Ronchetti M, Wein W, Navab N, Zettinig O, Prevost R. 2023.
Disa: Differentiable similarity approximation for universal multimodal registration.
\textit{MICCAI 2023}

\bibitem{mnih2015human}
Mnih V, Kavukcuoglu K, Silver D, Rusu AA, Veness J, et~al. 2015.
Human-level control through deep reinforcement learning.
\textit{nature} 518(7540):529--533

\bibitem{hase2020ultrasound}
Hase H, Azampour MF, Tirindelli M, Paschali M, Simson W, et~al. 2020.
\textit{Ultrasound-guided robotic navigation with deep reinforcement learning}.
In \textit{2020 IEEE/RSJ International Conference on Intelligent Robots and Systems (IROS)}, pp.  5534--5541. IEEE

\bibitem{li2021autonomous}
Li K, Wang J, Xu Y, Qin H, Liu D, et~al. 2021{\natexlab{b}}.
\textit{Autonomous navigation of an ultrasound probe towards standard scan planes with deep reinforcement learning}.
In \textit{2021 IEEE International Conference on Robotics and Automation (ICRA)}, pp.  8302--8308. IEEE

\bibitem{karamalis2012ultrasound}
Karamalis A, Wein W, Klein T, Navab N. 2012.
Ultrasound confidence maps using random walks.
\textit{Medical image analysis} 16(6):1101--1112

\bibitem{ning2021autonomic}
Ning G, Zhang X, Liao H. 2021.
Autonomic robotic ultrasound imaging system based on reinforcement learning.
\textit{IEEE Transactions on Biomedical Engineering} 68(9):2787--2797

\bibitem{bi2022vesnet}
Bi Y, Jiang Z, Gao Y, Wendler T, Karlas A, Navab N. 2022.
Vesnet-rl: Simulation-based reinforcement learning for real-world us probe navigation.
\textit{IEEE Robotics and Automation Letters} 7(3):6638--6645

\bibitem{li2023style}
Li K, Mao X, Ye C, Li A, Xu Y, Meng MQH. 2023.
Style transfer enabled sim2real framework for efficient learning of robotic ultrasound image analysis using simulated data.
\textit{arXiv preprint arXiv:2305.09169}

\bibitem{men2022multimodal}
Men Q, Teng C, Drukker L, Papageorghiou AT, Noble JA. 2022.
\textit{Multimodal-GuideNet: Gaze-Probe Bidirectional Guidance in Obstetric Ultrasound Scanning}.
In \textit{International Conference on Medical Image Computing and Computer-Assisted Intervention}, pp.  94--103. Springer

\bibitem{deng2021learning}
Deng X, Chen Y, Chen F, Li M. 2021.
\textit{Learning robotic ultrasound scanning skills via human demonstrations and guided explorations}.
In \textit{2021 IEEE International Conference on Robotics and Biomimetics (ROBIO)}, pp.  372--378. IEEE

\bibitem{abbeel2004apprenticeship}
Abbeel P, Ng AY. 2004.
\textit{Apprenticeship learning via inverse reinforcement learning}.
In \textit{Proceedings of the twenty-first international conference on Machine learning}, pp. ~1

\bibitem{ziebart2008maximum}
Ziebart BD, Maas AL, Bagnell JA, Dey AK, et~al. 2008.
\textit{Maximum entropy inverse reinforcement learning.}
In \textit{Aaai}, vol.~8, pp.  1433--1438. Chicago, IL, USA

\bibitem{burke2023learning}
Burke M, Lu K, Angelov D, Strai{\v{z}}ys A, Innes C, et~al. 2023.
Learning rewards from exploratory demonstrations using probabilistic temporal ranking.
\textit{Autonomous Robots} :1--19

\bibitem{d2006towards}
d’Aulignac D, Laugier C, Troccaz J, Vieira S. 2006.
Towards a realistic echographic simulator.
\textit{Medical image analysis} 10(1):71--81

\bibitem{goksel2009b}
Goksel O, Salcudean SE. 2009.
B-mode ultrasound image simulation in deformable 3-d medium.
\textit{IEEE transactions on medical imaging} 28(11):1657--1669

\bibitem{treeby2010k}
Treeby BE, Cox BT. 2010.
k-wave: Matlab toolbox for the simulation and reconstruction of photoacoustic wave fields.
\textit{Journal of biomedical optics} 15(2):021314--021314

\bibitem{gao2009fast}
Gao H, Choi HF, Claus P, Boonen S, Jaecques S, et~al. 2009.
A fast convolution-based methodology to simulate 2-dd/3-d cardiac ultrasound images.
\textit{IEEE transactions on ultrasonics, ferroelectrics, and frequency control} 56(2):404--409

\bibitem{burger2012real}
Burger B, Bettinghausen S, Radle M, Hesser J. 2012.
Real-time gpu-based ultrasound simulation using deformable mesh models.
\textit{IEEE transactions on medical imaging} 32(3):609--618

\bibitem{salehi2015patient}
Salehi M, Ahmadi SA, Prevost R, Navab N, Wein W. 2015.
\textit{Patient-specific 3D ultrasound simulation based on convolutional ray-tracing and appearance optimization}.
In \textit{Medical Image Computing and Computer-Assisted Intervention--MICCAI 2015: 18th International Conference, Munich, Germany, October 5-9, 2015, Proceedings, Part II 18}, pp.  510--518. Springer

\bibitem{mattausch2018realistic}
Mattausch O, Makhinya M, Goksel O. 2018.
\textit{Realistic Ultrasound Simulation of Complex Surface Models Using Interactive Monte-Carlo Path Tracing}.
In \textit{Computer Graphics Forum}, vol.~37, pp.  202--213. Wiley Online Library

\bibitem{tom2018simulating}
Tom F, Sheet D. 2018.
\textit{Simulating patho-realistic ultrasound images using deep generative networks with adversarial learning}.
In \textit{2018 IEEE 15th international symposium on biomedical imaging (ISBI 2018)}, pp.  1174--1177. IEEE

\bibitem{hu2017freehand}
Hu Y, Gibson E, Lee LL, Xie W, Barratt DC, et~al. 2017.
\textit{Freehand ultrasound image simulation with spatially-conditioned generative adversarial networks}.
In \textit{Molecular Imaging, Reconstruction and Analysis of Moving Body Organs, and Stroke Imaging and Treatment: Fifth International Workshop, CMMI 2017, Second International Workshop, RAMBO 2017, and First International Workshop, SWITCH 2017, Held in Conjunction with MICCAI 2017, Qu{\'e}bec City, QC, Canada, September 14, 2017, Proceedings 5}, pp.  105--115. Springer

\bibitem{vitale2020improving}
Vitale S, Orlando JI, Iarussi E, Larrabide I. 2020.
Improving realism in patient-specific abdominal ultrasound simulation using cyclegans.
\textit{International journal of computer assisted radiology and surgery} 15(2):183--192

\bibitem{zhang2020generalizing}
Zhang L, Wang X, Yang D, Sanford T, Harmon S, et~al. 2020.
Generalizing deep learning for medical image segmentation to unseen domains via deep stacked transformation.
\textit{IEEE Trans. Med. Imag.} 39(7):2531--2540

\bibitem{yin2020automatic}
Yin S, Peng Q, Li H, Zhang Z, You X, et~al. 2020.
Automatic kidney segmentation in ultrasound images using subsequent boundary distance regression and pixelwise classification networks.
\textit{Medical image analysis} 60:101602

\bibitem{lee2021principled}
Lee LH, Gao Y, Noble JA. 2021.
\textit{Principled ultrasound data augmentation for classification of standard planes}.
In \textit{IPMI}, pp.  729--741. Springer

\bibitem{pang2021semi}
Pang T, Wong JHD, Ng WL, Chan CS. 2021.
Semi-supervised gan-based radiomics model for data augmentation in breast ultrasound mass classification.
\textit{Computer Methods and Programs in Biomedicine} 203:106018

\bibitem{tiago2022data}
Tiago C, Gilbert A, Beela AS, Aase SA, Snare SR, et~al. 2022.
A data augmentation pipeline to generate synthetic labeled datasets of 3d echocardiography images using a gan.
\textit{IEEE Access} 10:98803--98815

\bibitem{shi2020knowledge}
Shi G, Wang J, Qiang Y, Yang X, Zhao J, et~al. 2020.
Knowledge-guided synthetic medical image adversarial augmentation for ultrasonography thyroid nodule classification.
\textit{Computer Methods and Programs in Biomedicine} 196:105611

\bibitem{zhang2021progressive}
Zhang R, Lu W, Wei X, Zhu J, Jiang H, et~al. 2021.
A progressive generative adversarial method for structurally inadequate medical image data augmentation.
\textit{IEEE Journal of Biomedical and Health Informatics} 26(1):7--16

\bibitem{pesteie2019adaptive}
Pesteie M, Abolmaesumi P, Rohling RN. 2019.
Adaptive augmentation of medical data using independently conditional variational auto-encoders.
\textit{IEEE transactions on medical imaging} 38(12):2807--2820

\bibitem{wulff2023towards}
Wulff D, Dohnke T, Nguyen NT, Ernst F. 2023.
\textit{Towards Realistic 3D Ultrasound Synthesis: Deformable Augmentation using Conditional Variational Autoencoders}.
In \textit{2023 IEEE 36th International Symposium on Computer-Based Medical Systems (CBMS)}, pp.  821--826. IEEE

\bibitem{tirindelli2021rethinking}
Tirindelli M, Eilers C, Simson W, Paschali M, Azampour MF, Navab N. 2021.
\textit{Rethinking ultrasound augmentation: A physics-inspired approach}.
In \textit{MICCAI}, pp.  690--700. Springer

\bibitem{yu2023mining}
Yu H, Li Y, Wu Q, Zhao Z, Chen D, et~al. 2023.
Mining negative temporal contexts for false positive suppression in real-time ultrasound lesion detection.
\textit{arXiv preprint arXiv:2305.18060}

\bibitem{painchaud2020cardiac}
Painchaud N, Skandarani Y, Judge T, Bernard O, Lalande A, Jodoin PM. 2020.
Cardiac segmentation with strong anatomical guarantees.
\textit{IEEE transactions on medical imaging} 39(11):3703--3713

\bibitem{jiang2023defcor}
Jiang Z, Zhou Y, Cao D, Navab N. 2023{\natexlab{d}}.
Defcor-net: Physics-aware ultrasound deformation correction.
\textit{Medical Image Analysis} :102923

\bibitem{jiang2021deformation}
Jiang Z, Zhou Y, Bi Y, Zhou M, Wendler T, Navab N. 2021{\natexlab{c}}.
Deformation-aware robotic 3d ultrasound.
\textit{IEEE Robotics and Automation Letters} 6(4):7675--7682

\bibitem{dou2022localizing}
Dou H, Han L, He Y, Xu J, Ravikumar N, et~al. 2022.
\textit{Localizing the Recurrent Laryngeal Nerve via Ultrasound with a Bayesian Shape Framework}.
In \textit{International Conference on Medical Image Computing and Computer-Assisted Intervention}, pp.  258--267. Springer

\bibitem{painchaud2022echocardiography}
Painchaud N, Duchateau N, Bernard O, Jodoin PM. 2022.
Echocardiography segmentation with enforced temporal consistency.
\textit{IEEE Transactions on Medical Imaging} 41(10):2867--2878

\bibitem{gare2022w}
Gare GR, Li J, Joshi R, Magar R, Vaze MP, et~al. 2022.
W-net: Dense and diagnostic semantic segmentation of subcutaneous and breast tissue in ultrasound images by incorporating ultrasound rf waveform data.
\textit{Medical Image Analysis} 76:102326

\bibitem{ning2023doppler}
Ning G, Liang H, Chen F, Zhang X, Liao H. 2023{\natexlab{b}}.
\textit{Doppler Image-Based Weakly-Supervised Vascular Ultrasound Segmentation with Transformer}.
In \textit{2023 IEEE 20th International Symposium on Biomedical Imaging (ISBI)}, pp.  1--5. IEEE

\bibitem{li2021image}
Li K, Xu Y, Wang J, Ni D, Liu L, Meng MQH. 2021{\natexlab{c}}.
Image-guided navigation of a robotic ultrasound probe for autonomous spinal sonography using a shadow-aware dual-agent framework.
\textit{IEEE Transactions on Medical Robotics and Bionics} 4(1):130--144

\bibitem{klein2015rf}
Klein T, Wells WM. 2015.
\textit{RF ultrasound distribution-based confidence maps}.
In \textit{Medical Image Computing and Computer-Assisted Intervention--MICCAI 2015: 18th International Conference, Munich, Germany, October 5-9, 2015, Proceedings, Part II 18}, pp.  595--602. Springer

\bibitem{meng2019weakly}
Meng Q, Sinclair M, Zimmer V, Hou B, Rajchl M, et~al. 2019{\natexlab{a}}.
Weakly supervised estimation of shadow confidence maps in fetal ultrasound imaging.
\textit{IEEE transactions on medical imaging} 38(12):2755--2767

\bibitem{mildenhall2021nerf}
Mildenhall B, Srinivasan PP, Tancik M, Barron JT, Ramamoorthi R, Ng R. 2021.
Nerf: Representing scenes as neural radiance fields for view synthesis.
\textit{Communications of the ACM} 65(1):99--106

\bibitem{wysocki2023ultra}
Wysocki M, Azampour MF, Eilers C, Busam B, Salehi M, Navab N. 2023.
Ultra-nerf: Neural radiance fields for ultrasound imaging.
\textit{MIDL 2023}

\bibitem{milletari2016v}
Milletari F, Navab N, Ahmadi SA. 2016.
\textit{V-net: Fully convolutional neural networks for volumetric medical image segmentation}.
In \textit{2016 fourth international conference on 3D vision (3DV)}, pp.  565--571. Ieee

\bibitem{degel2018domain}
Degel MA, Navab N, Albarqouni S. 2018.
\textit{Domain and geometry agnostic CNNs for left atrium segmentation in 3D ultrasound}.
In \textit{Medical Image Computing and Computer Assisted Intervention--MICCAI 2018: 21st International Conference, Granada, Spain, September 16-20, 2018, Proceedings, Part IV 11}, pp.  630--637. Springer

\bibitem{meng2019representation}
Meng Q, Pawlowski N, Rueckert D, Kainz B. 2019{\natexlab{b}}.
\textit{Representation disentanglement for multi-task learning with application to fetal ultrasound}.
In \textit{Smart Ultrasound Imaging and Perinatal, Preterm and Paediatric Image Analysis: First International Workshop, SUSI 2019, and 4th International Workshop, PIPPI 2019, Held in Conjunction with MICCAI 2019, Shenzhen, China, October 13 and 17, 2019, Proceedings 4}, pp.  47--55. Springer

\bibitem{meng2020unsupervised}
Meng Q, Rueckert D, Kainz B. 2020.
\textit{Unsupervised cross-domain image classification by distance metric guided feature alignment}.
In \textit{Medical Ultrasound, and Preterm, Perinatal and Paediatric Image Analysis: First International Workshop, ASMUS 2020, and 5th International Workshop, PIPPI 2020, Held in Conjunction with MICCAI 2020, Lima, Peru, October 4-8, 2020, Proceedings 1}, pp.  146--157. Springer

\bibitem{ying2022multi}
Ying X, Liu Z, Gao J, Zhang R, Jiang H, Wei X. 2022.
\textit{Multi-task Class Feature Space Fusion Domain Adaptation Network for Thyroid Ultrasound Images: Research on Generalization of Smart Healthcare Systems}.
In \textit{International Conference on Wireless Algorithms, Systems, and Applications}, pp.  139--152. Springer

\bibitem{morris2019weisfeiler}
Morris C, Ritzert M, Fey M, Hamilton WL, Lenssen JE, et~al. 2019.
\textit{Weisfeiler and leman go neural: Higher-order graph neural networks}.
In \textit{Proceedings of the AAAI conference on artificial intelligence}, vol.~33, pp.  4602--4609

\bibitem{zhang2023semi}
Zhang K, Li Z, Cai C, Liu J, Xu D, et~al. 2023{\natexlab{b}}.
Semi-supervised graph convolutional networks for the domain adaptive recognition of thyroid nodules in cross-device ultrasound images.
\textit{Medical Physics}

\bibitem{belghazi2018mutual}
Belghazi MI, Baratin A, Rajeshwar S, Ozair S, Bengio Y, et~al. 2018.
\textit{Mutual information neural estimation}.
In \textit{International conference on machine learning}, pp.  531--540. PMLR

\bibitem{cha2022domain}
Cha J, Lee K, Park S, Chun S. 2022.
\textit{Domain generalization by mutual-information regularization with pre-trained models}.
In \textit{European Conference on Computer Vision}, pp.  440--457. Springer

\bibitem{liu2021mutual}
Liu X, Yang C, You J, Kuo CCJ, Kumar BV. 2021.
Mutual information regularized feature-level frankenstein for discriminative recognition.
\textit{IEEE Transactions on Pattern Analysis and Machine Intelligence} 44(9):5243--5260

\bibitem{peng2019domain}
Peng X, Huang Z, Sun X, Saenko K. 2019.
\textit{Domain agnostic learning with disentangled representations}.
In \textit{International Conference on Machine Learning}, pp.  5102--5112. PMLR

\bibitem{meng2020mutual}
Meng Q, Matthew J, Zimmer VA, Gomez A, Lloyd DF, et~al. 2020.
Mutual information-based disentangled neural networks for classifying unseen categories in different domains: Application to fetal ultrasound imaging.
\textit{IEEE Trans. Med. Imag.} 40(2):722--734

\bibitem{bi2023mi}
Bi Y, Jiang Z, Clarenbach R, Ghotbi R, Karlas A, Navab N. 2023.
Mi-segnet: Mutual information-based us segmentation for unseen domain generalization.
\textit{MICCAI 2023}

\bibitem{fichtinger2022image}
Fichtinger G, Troccaz J, Haidegger T. 2022.
Image-guided interventional robotics: Lost in translation?
\textit{Proceedings of the IEEE} 110(7):932--950

\bibitem{khamis2019ai}
Khamis A, Li H, Prestes E, Haidegger T. 2019.
Ai: a key enabler of sustainable development goals, part 1 [industry activities].
\textit{IEEE Robotics \& Automation Magazine} 26(3):95--102

\bibitem{haidegger2019autonomy}
Haidegger T. 2019.
Autonomy for surgical robots: Concepts and paradigms.
\textit{IEEE Transactions on Medical Robotics and Bionics} 1(2):65--76

\bibitem{prestes2021first}
Prestes E, Houghtaling MA, Gon{\c{c}}alves PJ, Fabiano N, Ulgen O, et~al. 2021.
The first global ontological standard for ethically driven robotics and automation systems [standards].
\textit{IEEE Robotics \& Automation Magazine} 28(4):120--124

\bibitem{haidegger2022robot}
Haidegger T, Speidel S, Stoyanov D, Satava RM. 2022.
Robot-assisted minimally invasive surgery—surgical robotics in the data age.
\textit{Proceedings of the IEEE} 110(7):835--846

\bibitem{attanasio2021autonomy}
Attanasio A, Scaglioni B, De~Momi E, Fiorini P, Valdastri P. 2021.
Autonomy in surgical robotics.
\textit{Annual Review of Control, Robotics, and Autonomous Systems} 4:651--679

\bibitem{karlas2021optoacoustic}
Karlas A, Pleitez MA, Aguirre J, Ntziachristos V. 2021.
Optoacoustic imaging in endocrinology and metabolism.
\textit{Nature Reviews Endocrinology} 17(6):323--335

\bibitem{little2020optical}
Little C, Colchester R, Noimark S, Manmathan G, Rakhit R, Desjardins A. 2020.
Optical ultrasound (opus): a novel concept for intravascular imaging.
\textit{European Heart Journal} 41(Supplement\_2):ehaa946--2457

\bibitem{zhang2023miniaturised}
Zhang S, Lim CS, Zhang EZ, Beard PC, Desjardins AE, Colchester RJ. 2023{\natexlab{c}}.
\textit{Miniaturised all-optical ultrasound probe for thrombus imaging}.
In \textit{Opto-Acoustic Methods and Applications in Biophotonics VI}, vol. 12631, pp.  87--91. SPIE

\bibitem{graham2019vivo}
Graham M, Assis F, Allman D, Wiacek A, Gonzalez E, et~al. 2019.
In vivo demonstration of photoacoustic image guidance and robotic visual servoing for cardiac catheter-based interventions.
\textit{IEEE transactions on medical imaging} 39(4):1015--1029

\bibitem{shubert2017photoacoustic}
Shubert J, Bell MAL. 2017.
\textit{Photoacoustic based visual servoing of needle tips to improve biopsy on obese patients}.
In \textit{2017 IEEE International Ultrasonics Symposium (IUS)}, pp.  1--4. IEEE

\bibitem{hu2018stretchable}
Hu H, Zhu X, Wang C, Zhang L, Li X, et~al. 2018.
Stretchable ultrasonic transducer arrays for three-dimensional imaging on complex surfaces.
\textit{Science advances} 4(3):eaar3979

\bibitem{wang2021continuous}
Wang C, Qi B, Lin M, Zhang Z, Makihata M, et~al. 2021.
Continuous monitoring of deep-tissue haemodynamics with stretchable ultrasonic phased arrays.
\textit{Nature biomedical engineering} 5(7):749--758

\bibitem{wang2022bioadhesive}
Wang C, Chen X, Wang L, Makihata M, Liu HC, et~al. 2022{\natexlab{b}}.
Bioadhesive ultrasound for long-term continuous imaging of diverse organs.
\textit{Science} 377(6605):517--523

\bibitem{hu2023wearable}
Hu H, Huang H, Li M, Gao X, Yin L, et~al. 2023.
A wearable cardiac ultrasound imager.
\textit{Nature} 613(7945):667--675

\bibitem{lin2023fully}
Lin M, Zhang Z, Gao X, Bian Y, Wu RS, et~al. 2023.
A fully integrated wearable ultrasound system to monitor deep tissues in moving subjects.
\textit{Nature Biotechnology} :1--10

\end{thebibliography}

\end{document}